\documentclass[letterpaper, 10 pt, conference]{ieeeconf}  % Comment this line out if you need a4paper

\IEEEoverridecommandlockouts                              % This command is only needed if 
\usepackage{amsmath} % assumes amsmath package installed
\usepackage{amssymb}  % assumes amsmath package installed
\usepackage{graphicx}
\usepackage{xcolor}

\usepackage{algorithm}
\usepackage{algorithmic}
\usepackage{flushend}
\usepackage{makecell}
\usepackage{color}
\usepackage{overpic}
\usepackage{soul}
\usepackage{csquotes}
\usepackage{multirow}
\usepackage{multicol}
\usepackage{wrapfig}
\usepackage{booktabs}
\usepackage{threeparttable}
\usepackage{hyperref}

\newtheorem{theorem}{Theorem}
\newtheorem{definition}{Definition}

\title{\LARGE \bf
Distributed Multi-agent Interaction Generation \\with Imagined Potential Games
}

\author{Lingfeng Sun$^{1}$ \quad Pin-Yun Hung$^{1}$ \quad Changhao Wang$^{1}$ \quad Masayoshi Tomizuka$^{1}$ \quad Zhuo Xu$^{2}$% <-this % stops a space
\thanks{$^{1}$Department of Mechanical Engineering, 
        University of California, Berkeley, Berkeley, CA 94720, USA.
        {\tt\small \{lingfengsun, pinyun\_hung, changhaowang, tomizuka\}@berkeley.edu}}%
\thanks{$^{2}$Google Deepmind, Mountain View, CA 94043, USA.
        {\tt\small zhuoxu@google.com}}%
}

\begin{document}

\maketitle
\thispagestyle{empty}
\pagestyle{empty}

%%%%%%%%%%%%%%%%%%%%%%%%%%%%%%%%%%%%%%%%%%%%%%%%%%%%%%%%%%%%%%%%%%%%%%%%%%%%%%%%
\begin{abstract}
Interactive behavior modeling of multiple agents is an essential challenge in simulation, especially in scenarios when agents need to avoid collisions and cooperate at the same time. 
Humans can interact with others without explicit communication and navigate in scenarios when cooperation is required.
In this work, we aim to model human interactions in this realistic setting, where each agent acts based on its observation and does not communicate with others.
We propose a framework based on distributed potential games, where each agent imagines a cooperative game with other agents and solves the game using its estimation of their behavior. We utilize iLQR to solve the games and closed-loop simulate the interactions. 
We demonstrate the benefits of utilizing distributed imagined games in our framework through various simulation experiments. We show the high success rate, the increased navigation efficiency, and the ability to generate rich and realistic interactions with interpretable parameters. Illustrative examples are available at \href{https://sites.google.com/berkeley.edu/distributed-interaction}{https://sites.google.com/berkeley.edu/distributed-interaction}.
\end{abstract}

%%%%%%%%%%%%%%%%%%%%%%%%%%%%%%%%%%%%%%%%%%%%%%%%%%%%%%%%%%%%%%%%%%%%%%%%%%%%%%%%
\section{INTRODUCTION}
Modeling the interactive behavior of multiple agents in different scenarios is an essential task in crowd simulation. One of the main challenges is to model the interactive behaviors of multiple agents, especially in narrow scenarios where they have to avoid obstacles at the same time.
The agents' decisions are interdependent, meaning that each agent’s decision influences and is influenced by the decision of the other agents. %, and the closed-loop behaviors are affected by the predictions of other agents.
In real life, humans can cooperate with each other without explicit communication; instead, we make decisions based on observations. In this work, we aim to analyze and model the interactions in a distributed game setting without communication.

\begin{figure}[tb]
\begin{center}
	\includegraphics[width=0.5\textwidth]{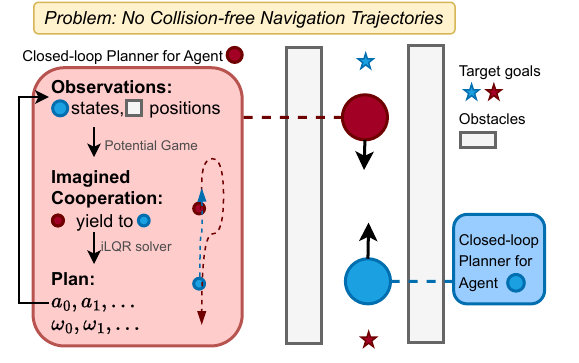}
	\caption{A narrow-way problem challenging to solve in the distributed and no-communication setting. There's no collision-free for single-agent navigation, and agents must cooperate (one moving backward to yield the other agent) to solve the problem. We propose adding imagined cooperation in distributed planning to simulate cooperative interactions.}
	\label{fig: narrow_infeasible_case}
\end{center}
\vspace{-0.25in}
\end{figure}

The problems can be naturally formulated as a multi-agent planning problem with separate goals and a shared environment.
Some previous works have used centralized~\cite{2016tang,2020michael} algorithms, which solve trajectories of all agents together to control all robots. However, centralized methods are computationally expensive and require full information about the environment and other agents. In contrast, distributed algorithms separately solve the short-horizon reaction plans~\cite{orca,bvc} or long-horizon trajectory plans~\cite{rlss,dmpc,wang2021trajectory} for each robot. Distributed methods are more scalable and robust but suffer from deadlocks, especially in human-like interaction cases where no communication is allowed. For instance, as shown in Figure \ref{fig: narrow_infeasible_case}, in a narrow-way situation, if both agents are in the hallway, they have no collision-free navigation plans to their goals. If agents have no information from the others, they need to estimate other's intentions and figure out how to cooperate in the distributed setting.

We propose to model and solve the interaction problem in a distributed manner where each agent imagines a cooperation game with others.
There are several works considering game-theoretic frameworks~\cite{2018potentialgame,2019fisac,pilqr,2019wang} to model cooperative behaviors. We follow the dynamic game formulation introduced by \cite{pilqr}, where the multi-agent potential game is formulated into an equivalent single optimal control problem to find cooperative plans for all agents. We assume an ``imagined'' game exists in all agents' distributed planners to predict others' behavior and use the iterative LQR (iLQR) to solve optimal plans. In addition to the basic formulation in \cite{pilqr}, we add collision avoidance to environmental obstacles, observation range, and blind area for agents in the optimization since they are essential causes of human-like interactions. 

\begin{figure*}[t!]
    \centering
    \includegraphics[width=\textwidth]{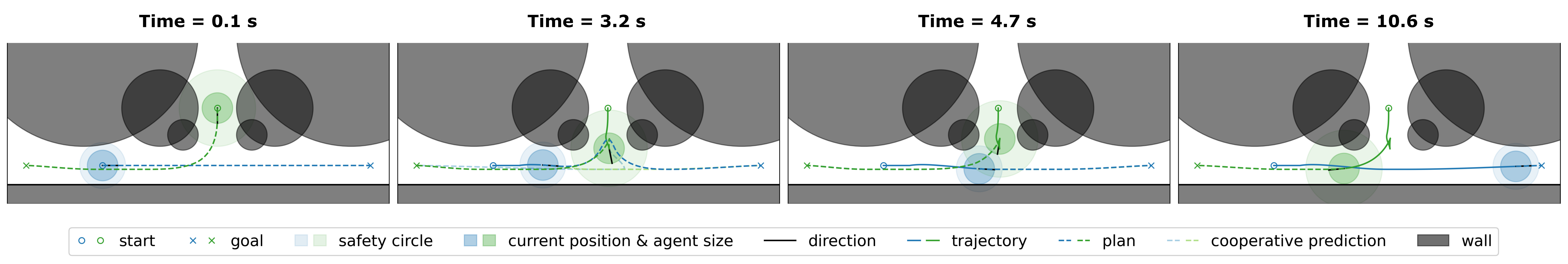}
    \caption{An example interaction generated in T-intersection. From left to right is the closed-loop trajectory over time. Solid lines are past trajectories; darker and lighter dotted lines are plans of ego agents and predictions of other agents (e.g., the light blue line is the predicted \textit{green} agent behavior by the \textit{blue} agent). All green lines are from the green agent's planner.}
    \label{fig:T-intersection}
\vspace{-0.15in}
\end{figure*}

We demonstrate the effectiveness of introducing the ``Imagined Potential Game (IPG)'' formulation into the multi-agent planning problem without communication. The open-loop plans are used to simulate the closed-loop behaviors using receding horizon control.
Agents may have different open-loop cooperative strategies but can gradually converge to cooperative behaviors in closed-loop simulation.
Experiments on the narrow-way scenario with random initialization empirically show the improved success rate and navigation efficiency in simulating cooperative interactions. Furthermore, we show the framework's capability to generate diverse and realistic interaction behaviors by varying parameters, such as safety distance or objective weights. Lastly, we demonstrate that the IPG-controlled agents can robustly interact with non-IPG-controller agents.
The contribution of this work lies in the following aspects:
\begin{enumerate}
    \item We propose a multi-agent interaction generation framework using the imagined potential games under a distributed and no-communication setting to simulate interactions in dense obstacle scenarios.
    \item We use simulated experiments to empirically demonstrate the improvement in success rate and navigation efficiency for simulating complex interactions.
    \item We demonstrate the algorithm can generate diverse and realistic interactions using interpretable parameters and interact with heterogeneous agents.
\end{enumerate}

% The rest of the paper is organized as follows: Sec II introduces the previous related works on interactive trajectory generation. Section III introduces our complete interaction generation framework. Section IV and V demonstrates the simulated results of various interactions and our discussions on the current framework.
% \zhuo{Like said, we would want to focus on telling one story. In this case, I think limiting the scope to the gametheory based interaction handling is good enough in terms of the contribution scope.}
% \zhuo{If you want to include the efforts of the MPC system setup, I'd suggest making it the secondary contribution, and use 70\% of the space talking about the game-based method and 30\% for the d-MPC}

\section{Related Works}
\subsection{Multi-robot planning}
Multi-robot trajectory planning algorithms can be categorized based on where the computation is done. Two main strategies to solve the problem are \textit{centralized} and \textit{distributed}. Centralized planning algorithms solve the trajectories for all the agents in the same problem utilizing the global information and then send the control commands to the robots.
Most previous works use multi-agent path finding (MAPF) solvers with trajectory optimization algorithms\cite{2016tang, 2020park, 2020michael, wang2022bpomp} to find feasible trajectories for all agents.
A series of works \cite{2019fisac,2020spica, 2019wang, 2022miller, 2022algames, 2023laine, 2023mehr} models interactions in multi-agent planning via game theoretic frameworks. \cite{pilqr} utilizes potential games~\cite{2018potentialgame} for multi-agent trajectory planning with symmetric inter-agent costs.
Centralized algorithms can provide theoretical guarantees for planning success and optimality. 
However, the centralized setting cannot simulate realistic interactive behaviors in the distributed setting, where single-agent behaviors are affected by other agents' online reactions.

In the distributed setting, each agent runs a separate algorithm to compute its own trajectory. Depending on whether communication exists between the agents. 
Reactive algorithms\cite{orca, 2017wang, 2023jian} like Optimal Reciprocal Collision Avoidance (ORCA) can effectively avoid collisions but fail to avoid deadlocks in environments with dense obstacles\cite{rlss}. Learning-based reactive strategies\cite{primal, glas, gnn, batra2021} are computationally more efficient but suffer from distribution shifts and also experience deadlocks.
Another series of works like dMPC\cite{dmpc} and MADER\cite{mader} consider longer horizons and generate sequences instead of single actions; however, they require communication of plans for collision avoidance.
\cite{distributed_pg} turns multiple agents into distributed groups and solves the in-group interactions, but the in-group agents are modeled in the centralized setting.  
RLSS\cite{rlss} uses a fully distributed setting and requires only the current state sensing of other agents to plan piece-wise Bézier trajectories to improve deadlock performance.
However, these methods don't explicitly model the cooperation of agents and, therefore, have trouble simulating interactions when the initial condition is not feasible for collision-free planning. 

\subsection{Interaction Generation}
Simulating the behaviors of actors is an important task with a wide range of applications in transportation and robotics research since simulators play essential roles in training and evaluating intelligent agents like indoor robots and autonomous vehicles.
Many autonomous-driving research works focus on simulating distributed agents' behaviors in the simulator to generate realistic interactions and reactive agents\cite{Yeh_2019_CVPR, Suo_2021_CVPR, 2021diverse, 2023editing, 2022trajgen} or analyzing and predicting interactive behaviors of heterogeneous agents\cite{qcnet, 2022pseudo, 2019itsc}.
Previous works on crowd simulation\cite{crowd_Piccoli_2009} focus more on the scalability of simulating agents\cite{Toll2015, roland2012}, and grouping in the crowd\cite{2009crowdgroup, mahato2017particle}. In this paper, we focus on the interactions in scenarios where cooperation is required to simulate multi-agent behaviors.

\section{Preliminaries}
\subsection{Distributed multi-agent planning}
\label{sec:distributed_problem}
Assume we have $N$ agents in the scenario. For each agent $i, 1\leq i\leq N$, let the vector $x_i(t)\in \mathbb{R}^{n_i}$  denote the state of agent $i$, let $u_i(t)\in \mathbb{R}^{m_i}$ denote the control input of agent $i$ at time $t$. Each agent follows its system dynamics 

$$x_i(k+1)=f_i(x_i(k), u_i(k))$$
Unless otherwise specified, throughout this paper, for variable $x$, we use a subscript \(x_i\) to denote agent $i$ and a superscript \(x^k\) or \(x(k)\) to indicate the time horizon $k$. If $i, k$ are not specified, it means $x$ for all agents or across all time steps. $x_{-i}$ denotes $x$ of all agents excluding agent $i$. 

Obstacles in the scenario are represented by $\{O_j\}_{j=1}^M$. Each agent has its initial state $x_i^0$ and a target goal state in the scenario $g_i$. All the agents in the scenario navigate to their target goal while avoiding collision with the environment and other agents. Interactions happen when their planned trajectories $\{x_i(0), x_i(1),...,x_i(T)\}_{i=1}^N$ have conflicts and need to interact to reach non-conflict new plans. Control inputs of multiple agents $U=[u_1^{0:T},u_2^{0:T},...,u_N^{0:T}]$ are the plans of all the agents.
Under the \textit{distributed setting with no communication}, we assume each agent $i$ is solving an optimal control optimization without knowing others' plans.
\begin{equation}
\begin{aligned}
\min_{u_i(0:T), x_i(0:T)} \quad & J_i(x(0), u_i,\Tilde{u}_{-i})\\
\textrm{s.t.} \quad & 
x_i(k+1) = f_i(x_i(k), u_i(k)) \\
&h(x_i,\Tilde{x}_{-i}, O) \leq 0\\
\end{aligned}
\label{eq:distributed_planning}
\end{equation}

$J_i=S_i(x(T), T) + \sum_{k=0}^{T-1}L_i(x(k), \Tilde{u}_{-i})$ is the cost function for agent $i$. The stage cost $L_i$ can include distance, time, and energy costs, and the terminal cost $S_i$ can include goal conditions. The collision-free requirements are described using the constraints $h\leq 0$. The hard constraints in $h$ can be added as weighted cost functions depending on the solver used. $\Tilde{u}_{-i}$ and $\Tilde{x}_{-i}$ are the estimation of other agents' plans and states to prevent collisions since we assume no communication between agents. To consider other agents during planning, it needs to use predictions. In most cases where environmental constraints are not strict, constant velocity predictions are well enough to provide collision avoidance planning. However, cooperative predictions are required to generate feasible interactions in cases like narrow-way interaction in Figure \ref{fig: narrow_infeasible_case}. Therefore, we proposed to use an Imagined Potential Game (IPG) framework to predict $\Tilde{x}_{-i}$ during the planning of agent $i$.

\textbf{Comparison with the \textit{centralized} or \textit{distributed with sharing} setting:} 
% \zhuo{you may want to add the p[timization formulation of the centralized planning here} 
In the \textit{centralized} setting, $U$ is solved together in a single large problem given all agents' initial and goal states simultaneously. The weighted ($\alpha_i$) costs of all agents are optimized in one problem.
\begin{equation}
\begin{aligned}
\min_{U} \quad & \sum_{i=1}^{N}\alpha_i J_i(x(0), U)\\
\textrm{s.t.} \quad & 
x(k+1) = f(x(k), u(k)), \quad h(x, O) \leq 0\\
\end{aligned}
\label{eqn_ori_formulation}
\end{equation}
The \textit{distributed setting with sharing} is quite similar to the centralized setting; plans are solved separately but are shared with other agents, enabling accurate predictions in a distributed setting \cite{dmpc} for cooperation. Many game-theoretical interaction models operate in a similar setting. The cost functions of all agents are shared to consider others' behaviors and find equilibrium plans for all agents. 

\subsection{Differentiable Potential Game}
\label{sec:potential_game}
We then introduce the definitions of dynamic games from previous works~\cite{pilqr}. We describe a differential game by the compact notation of $\Gamma_{x_0}^T=(N, \{U_i\}_{i=1}^N, \{J_i\}_{i=1}^N, \{f_i\}_{i=1}^N)$, where $x_0$ is the initial states of all agents, and each agent seeks to optimize its cost $J_i$ under the dynamic $f_i$. The cost function $J_i(x(0), U)=S_i(x(T)) + \sum_{k=0}^{T-1}L_i(x(k), U(k))$ consists of running cost $L_i$ and terminal cost $S_i$. We look for Nash equilibrium solution of the dynamic game defined by:
\begin{definition}
Given a differential game $\Gamma_{x_0}^T=(N, \{U_i\}_{i=1}^N, \{J_i\}_{i=1}^N, \{f_i\}_{i=1}^N)$, a set of control signals $U$ is an open-loop Nash equilibrium if, for every agent $i\in [N]$:
\begin{equation}
    J_i(x_0, u^*) \leq J_i(x_0, u_i, u_{-i}^*)
\end{equation}
\end{definition}

At a Nash equilibrium, no agent has the incentive to change its current control input $u_i^*$ as such a change would not yield any benefits, given that all other agents' controls $u_{-i}^*$ remain fixed. While this equilibrium solution can best represent the cooperative multi-agent behavior in the interaction, finding the Nash equilibrium solution is challenging since there are $N$ coupled optimal control problems to solve simultaneously. Recent progress in solving this problem, especially in robotics applications, find efficient solutions to the problems under certain conditions. As introduced in \cite{pilqr}, problems in a potential differential game form can be solved by formulating a single centralized optimal control problem. We summarize the result in the following theorem.

\begin{theorem}
\label{thm:pg}
For a given differential game $\Gamma_{x_0}^T=(N, \{U_i\}_{i=1}^N, \{J_i\}_{i=1}^N, \{f_i\}_{i=1}^N)$, if for each agent $i$, the running cost and terminal cost functions have the structure of
\begin{equation}
    L_i(x(k), u(k))=p(x(k), u(k))+c_i(x_{-i}(k), u_{-i}(k))
\end{equation}
\begin{equation}
    S_i(x(T) = \Bar{s}(x(T)) + s_i(x_{-i}(T))
\end{equation}
then the open-loop Nash equilibria can be found by solving the following optimal control problem
\begin{equation}
\begin{aligned}
\min_{U} \quad & \sum_{k=1}^{T-1}p(x(k), u(k))+\Bar{s}(x(T))\\
\textrm{s.t.} \quad & 
x_i(k+1) = f_i(x_i(k), u_i(k)) \\
\end{aligned}
\label{eq:pgeq}
\end{equation}
\end{theorem}

The key takeaway from this theorem is that one can formulate a differentiable potential game if all the cost function terms can be decomposed into potential functions ($p(\cdot), \Bar{s}(\cdot)$) that depend on the full state and control vectors of all the agents, and other cost terms ($c_i(\cdot), s_i(\cdot)$) that have no dependence on the state and control input of agent $i$. Then the optimal solution for both agents can be solved by the centralized problem Eq.\ref{eq:pgeq} using only $p, \Bar{s}$.
% \zhuo{and then omitting them? You should point out that to avoid confusions.}
The following section will show how this theorem is used for distributed no-communication settings. For each agent in the interaction, it assumes all agents are in the potential game, but it doesn't know the interaction parameters for other agents. Therefore, each agent will solve a separate Imagined Potential Game (IPG) using estimated parameters.

\subsection{System Notations and Assumptions}
To simplify the problem, we make several assumptions on system dynamics. 
We assume that all agents are modeled using the same unicycle dynamic model. The state vector $x_i = [p_{x, i}, p_{y, i}, \theta_i, v_i]^T$, the control vector $u_i = [a_i, w_i]^T$. 
% \zhuo{Also, shouldn't this be introduced earlier -- in the problem statement section?} 
The discrete-time dynamic equations of the system are: 

\begin{equation}
\begin{aligned}
    % x_i(k+1) &= f(x_i(k), u_i(k)) \\
    %          &= \begin{bmatrix}p_{x,i}(k+1) \\ p_{y,i}(k+1) \\ \theta_i(k+1) \\ v_i(k+1)\\ \end{bmatrix}\\
    %          &\\
    p_{x,i}(k+1) &= p_{x,i}(k) + T_s \ v_i(k) \ cos(\theta_i(k)) \\ 
    p_{y,i}(k+1) &= p_{y,i}(k) + T_s \ v_i(k) \ sin(\theta_i(k)) \\ 
    \theta_i(k+1) &= \theta_i(k) + T_s \ w_i(k) \\ 
    v_i(k+1) &= v_i(k) + T_s \ a_i(k)
\end{aligned}
\end{equation}

System notations are summarized in Table \ref{tab: notation}.

\begin{table}[h]
\centering
\caption{Notation describing common variable.}
\label{tab: notation}
\begin{tabular}{l l || l l }
    \Xhline{1pt}
     & Definition & & Definition (Default value)\\
    \hline
     $p_x, p_y$ & position in 2D & Q & state weight ([0.01, 0.01, 0, 0])\\
     $\theta$ & heading angle & R & input weight ([1, 1])\\
     v & velocity & D & safety weight (40) \\
     a & acceleration & B & back up weight (10)\\
     w & angular velocity & r & safety radius (1.2$\sim$2.0)\\
     $O$ & static obstacles & $T_s$ & sampling time (0.1) \\
    \Xhline{1pt}
\vspace{-0.15in}

\end{tabular}
\vspace{-0.1in}

\end{table}
\section{Distributed multi-agent interaction modeling with imagined potential game}
As described in section \ref{sec:distributed_problem}, for interactions under the distributed setting with no shared plans, each agent is solving the problem in Eq. \ref{eq:distributed_planning} but needs to estimate the future inputs and states of other agents $(\Tilde{x}_{-i}, \Tilde{u}_{-i})$ to avoid collision. One common assumption in navigation is the constant velocity prediction, which estimates other agents' states with the current velocity. However, this doesn't work for cooperation-required cases, as shown in Figure \ref{fig: narrow_infeasible_case}. The potential game formulation in section \ref{sec:potential_game} gives optimal actions for cooperative agents but assumes the cost parameters in $J_i$ are pre-known. In this paper, we use this formulation to make cooperative predictions of other agents with an Imagined Potential Game (IPG) setting but use estimated parameters of other agents to solve the game.

Parameters in agent $i$'s cost function are the goal position $g_i$ and the interaction parameters, including safety radius $r_i$, and different weights $Q_i, R_i, D_i, B_i$, each will affect the behavior in interaction. 
In this project, we assume other agents' long-term goal positions (intentions) are already predicted. For interaction parameters, agents will assume other agents using the same parameters they have, e.g., $\Tilde{r}_j=r_i$ for all $j\neq i$. This assumption is simple, but in practice, we found it strong enough to simulate diverse interactions, especially when all participants are cooperative agents. For cases where inaccurate estimation of others' parameters causes failure, we can update the estimation online (see section \ref{sec:heterogeneous}).

\subsection{Solving Potential game with estimated parameters}

Based on Theorem \ref{thm:pg}, Nash equilibrium solutions of all agents can be solved by Eq. \ref{eq:pgeq} if the interaction can be formulated into a potential game. Here, we show the running cost function $L_i(x)$ in the potential game, consisting of the stage cost term  $C_{tr,i}^{0:T-1}(x_i, u_i)$,  the collision avoidance term $C_{a,ij}(x_i, x_j)$ and the reverse avoidance term $C_{b,i}(x_i)$. The stage cost includes the minimum goal distance and input penalty terms:
\begin{equation}
\begin{aligned} 
C_{tr,i}^{0:T-1}(x_i, u_i) = (x_i - g_i)^TQ_i(x_i - g_i) + u_i^T R_i u_i
\end{aligned}
\end{equation}
The collision avoidance term is counted when the distance between two agents $d_{ij}$ is smaller than the safety distance:
\begin{equation}
    C_{a, ij}(x_i, x_j) =  \begin{cases} (d_{ij} - d_{safe})^2 \cdot D_i, & \text{if $d_{ij} < d_{safe}$} \\ 0, & \text{others} \end{cases}
\end{equation}
and satisfy the symmetric property $C_{a, ij}(x_i, x_j) = C_{a, ji}(x_j, x_i)$ for potential game described in \cite{pilqr}.

The reverse avoidance term discourages the agent for moving backward:

\begin{equation}
C_{b, i}(x_i) =  \begin{cases} |v_i| \cdot B_i, & \text{if $v_i < 0$} \\ 0, & \text{others} \end{cases}
\end{equation}

% Composing these three cost terms, the running cost function of agent $i$ can be represented as following : 

% \begin{equation}
% \begin{aligned} 
% L_i(x_i, u_i) = C_{tr,i}^{0:T-1}(x_i, u_i) + \sum_{i \neq j}^N C_{a, ij}(x_i, x_j) + C_{b, i}(x_i)
% \end{aligned}
% \end{equation}

To prove this running cost function is a potential game, we can represent the $p(x, u)$ and $c_i(x_{-i}, u_{-i})$ in Theorem \ref{thm:pg}:
\begin{equation}
\begin{aligned} 
p(x, u) = &\sum_{i=1}^N C_{tr, i}^{0:T-1}(x_i, u_i) \\ 
& + \sum_{1 \leq i < j} C_{a, ij}(x_i, x_j) + \sum_{i=1}^N C_{b, i}(x_i)\\
\end{aligned}
\end{equation}
\begin{equation}
\begin{aligned} 
c_i(x_{-i}, u_{-i}) = 
&-\sum_{j \neq i} C_{tr, j}^{0:T-1}(x_j, u_j) \\
&- \sum_{\substack{1 \leq j < k \\ j, k \neq i}} C_{a, jk}(x_j, x_k) - \sum_{j \neq i} C_{b, j}(x_j)
\end{aligned}
\end{equation}
With this representation, we show that the running cost $L_i(x_i, u_i)= p(x, u) + c_i(x_{-i}, u_{-i})$ follows the Theorem \ref{thm:pg}, 

% \begin{equation}
% \begin{aligned} 
% L_i(x_i, u_i) \
%     & = C_{tr,i}^{0:T-1}(x_i, u_i) + \sum_{i \neq j}^N C_{a, ij}(x_i, x_j) + C_{b, i}(x_i) \\
%     & = p(x, u) + c_i(x_{-i}, u_{-i})
% \end{aligned}
% \end{equation}
Similarly, the terminal cost $S_i(x(T)) = C_{tr,i}^T(x_i, u_i) = \bar{s}(x(T)) + s_i(x_{-i}(T))$, with terminal cost term:
\begin{equation}
\begin{aligned} 
C_{tr,i}^T(x_i, u_i) = (x_i(T) - g_i)^TQ_i(x_i(T) - g_i)
\end{aligned}
\end{equation}
Set $\bar{s}(x(T))$ and $s_i(x_{-i}(T))$ to be :
\begin{equation}
\begin{aligned} 
&\bar{s}(x^T) = \sum_{i=1}^N C_{tr, i}^T(x_i),s_i(x_{-i}^T) = -\sum_{j \neq i}^N C_{tr, j}^T(x_j)
\end{aligned}
\end{equation}
We have the required terminal cost $S_i(x(T))$ in Theorem \ref{thm:pg}. 

% \begin{equation}
% \begin{aligned} 
% S_i(x(T)) = C_{tr,i}^T(x_i, u_i) = \bar{s}(x(T)) + s_i(x_{-i}(T))
% \end{aligned}
% \end{equation}

% \zhuo{ You should describe what the gains stand for and in the experiment section give the magnitudes of your parameters} 

To address the problem in scenarios involving obstacles, we introduce some extra constraints in addition to the existing dynamic constraints, including the state boundary constraint, input constraint, and obstacle avoidance constraint. These constraints can be added as weighted costs into $J_i$ and don't affect the potential game assumptions. 

One important difference between centralized planning and the IPG setting is the safety distance $d_{safe}$. For centralized planning, the maximum safety distance between them is used $d_{safe} = max(r_i, r_j)$, for IPG, each agent assumes others have the same safety distance, $d_{safe, i}=r_i$, unless it changes its estimation. 

The full IPG problem for each agent to solve is:
\begin{equation}
\begin{aligned} 
    \min_{U}  \ &\sum_{k=1}^{T-1} p(x(k), u(k)) + \bar{s}(x(T))\\ 
    s.t. \quad &x(k+1) = f(x(k), u(k)) \\ 
    &x(0) = x_0 \\ 
    &x_L \leq x(k) \leq x_U \\ 
    &u_L \leq u(k) \leq u_U \\ 
    &r_{obs} - dis(x(k), O_m) \leq 0, m=1...M \\ 
    &r_{i} - dis(x_i(k), x_j(k)) \leq 0, i, j=1...N\\
\end{aligned}
\end{equation}
where $x_U, x_L, u_U, u_L$ are the state and input boundaries, and $r_{obs}$ is the radius of the circle obstacle.

\textbf{Algorithm 1} outlines the process of closed-loop simulation of the interaction: each agent solves the IPG using iLQR. Following the receding horizon control manner, only the first step for itself, denoted as $u_i^0$ is used. All agents can solve their IPG in parallel; the closed-loop simulating process continues until all agents are sufficiently close to their target goals. The stored trajectories $\{\tau_i\}_{i=0}^N$ are the generated interactions.

\begin{minipage}{0.465\textwidth}
% \vspace{-0.28 in}
        \begin{algorithm}[H]
            \centering
            \caption{Closed-loop Distributed multi-agent Planning using IPG}
            \label{alg:algo1}
                    \begin{algorithmic}
                        \STATE {\bfseries Initialize:} U = 0, $\{\tau_i\}_{i=0}^N$ = empty \\
                        \WHILE{termination condition not satisfied}
                         \STATE {\bfseries for} i in N {\bfseries do in parallel} \\
                         \STATE $\quad U \leftarrow iLQR(x_0, g, U, O, Q_i, R_i, D_i, B_i, d_{safe, i})$ \\
                         \STATE $\quad u_i^0 \leftarrow U=[u_1^{0:T},u_2^{0:T},...,u_N^{0:T}]$ \\
                         \STATE $\quad x_{next, i} \leftarrow f(x_0, u_i^0)$ \\
                         \STATE {\bfseries end for} \\

                          \STATE {\bfseries for} i in N {\bfseries do} \\
                         \STATE $\quad x_{0, i} \leftarrow x_{next, i}$\\
                         \STATE $\quad \tau_i \leftarrow [\tau_i, x_{0, i}]$  \\
                         \STATE {\bfseries end for} \\
                         \STATE $x_{0} \leftarrow x_{0, i=1,...,N}$\\
                         
                        % \STATE 	($\w$-reset) \, $\w_{\rm new} \leftarrow  \text{Eqn.(\ref{eq:reset})} \quad \text{if} \, \mathcal{L}_{\rm new} > \epsilon\, $\\
                        \ENDWHILE
                    \end{algorithmic}
        \end{algorithm}
\end{minipage}
\section{Experiments}
\begin{figure*}[t]
    \centering
    \includegraphics[width=\textwidth]{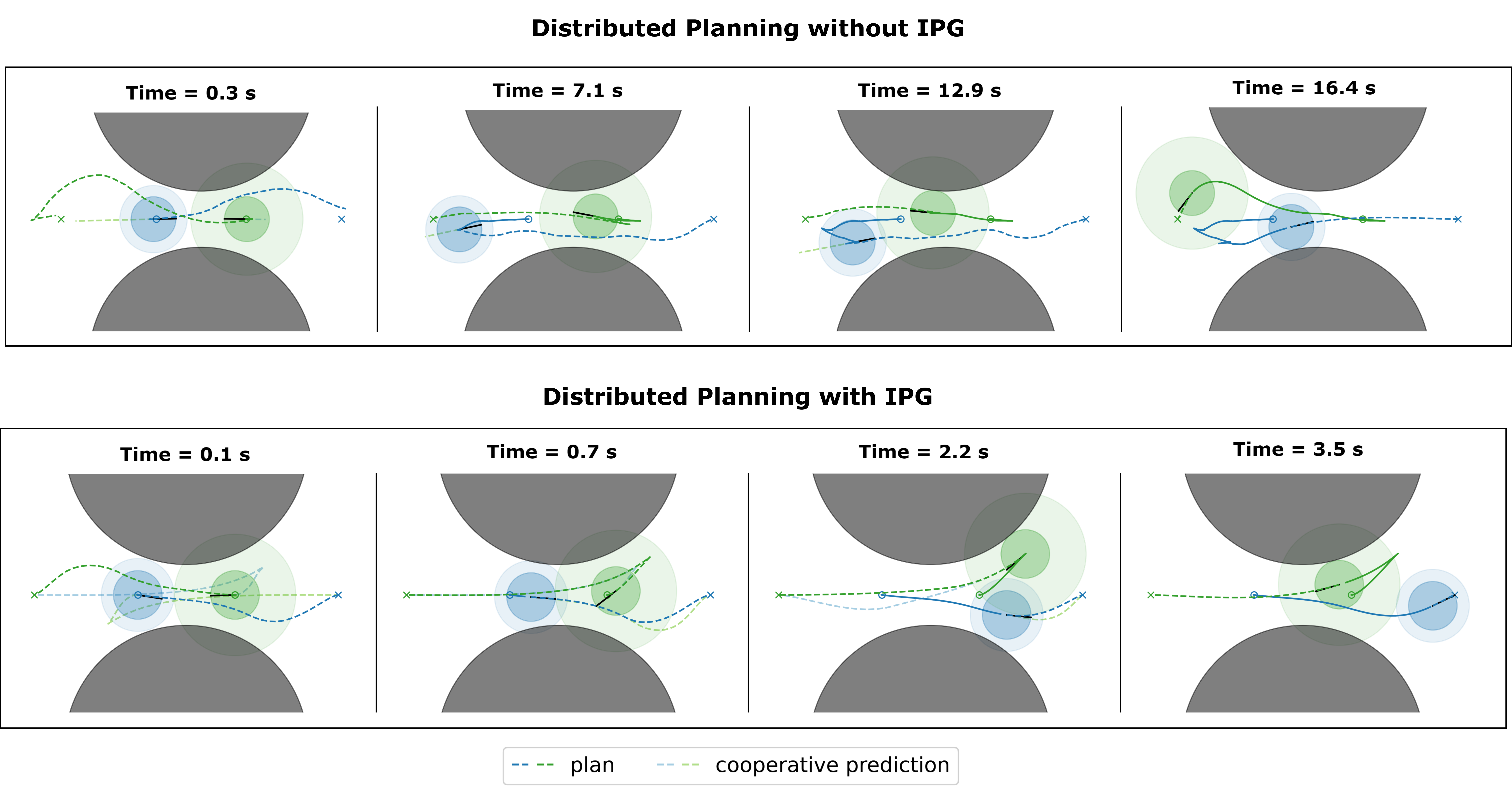}
    \vspace{-0.2in}
    \caption{In the distributed setting, Vanilla agents without IPG are stuck at a deadlock for a long time and take extremely long time to finish interaction. IPG agents cooperatively interact with each other.}
    \label{fig:distributed comparison}
\vspace{-0.18in}
\end{figure*}

In this section, we present the simulating results under the distributed multi-agent interaction setting to address the following key questions of interest: 
\begin{enumerate}
    \item Is the proposed framework able to solve the complex (infeasible) cases in navigation under the distributed setting? Does it prevent the failures that result from deadlock and collision?
    \item Is the proposed framework able to generate rich interactions with various parameters? Is the behavior realistic and diverse for crowd simulation?
\end{enumerate}

\noindent\textbf{Environment settings:} 
We first quantitatively evaluate and demonstrate the interaction generation capability in the classic ``narrow way'' problem, the most common but challenging indoor navigation scenario where only one agent can pass at one time. Then, we demonstrate and analyze the interactions with more agents or in more complex scenarios.

\noindent\textbf{Comparison baselines:} 
 We use two baselines to compare the effect of our proposed framework. 1) \textit{Vanilla:} This is the vanilla version of distributed multi-agent planning. Each agent is running a collision avoidance planning algorithm. The agent will follow the previous plan if the planner cannot generate a feasible collision-free solution. Note that an open-loop collision solution might not cause an actual collision in the closed-loop simulation since distributed agents use wrong predictions during open-loop planning. 2) \textit{Brake:} To avoid collision under infeasible solutions, the agent will brake and stop if the planner solution has collision. This conservative implementation can avoid collisions in simulation but is more likely to cause deadlocks. 

\noindent\textbf{Evaluation metrics:} 
We generate 20 test cases with randomized starting and goal points. To make sure interactions happen between agents, we enforce starting and goal points on different sides of the hallway. The safety and navigation parameters are randomly sampled from a pre-defined range for different agents.
The metrics for evaluating the interaction generation are 1) \textbf{\textit{Success rate}}: We test whether the two agents reach the target goals without collision before the time limit. For failure cases, we separate the failure caused by collision and failure by deadlocks. 2) \textbf{\textit{Extra interacting time}}: We use the total time for agents to finish interaction under a centralized setting as a baseline to test how much extra time the agents spent in the distributed setting.

\noindent\textbf{Observation range in distributed setting}: In the distributed setting, the ego-agent will consider other agents' behaviors unless they are blinded (e.g., there are obstacles on the line connecting the two agents) or other agents are behind the ego-agent and out of a pre-defined sensing range.

\subsection{Interaction in ``narrow way'' in distributed setting}
Both agents aim to reach the target positions sampled on the opposite side while avoiding collisions with other agents and the walls. 
The experiment results are shown in Table \ref{tab:narrowway}. 
\begin{table}[h]
    \begin{threeparttable}[b]
    \caption{Evaluation in Narrow-way Scenario}
    \vspace{-0.1in}
    \centering
    \begin{tabular}{l|lll|l}
\Xhline{1pt}
               & Success Rate & Deadlock & Collision & AET(second)  \\ 
    \hline
    Vanilla    & 13/20  & 2/20 &  5/20 & +2.054\\
    Brake      & 12/20  & 8/20 &  0/20 & +2.308   \\
    IPG(Ours)  & 20/20 &  0/20 &  0/20 & +0.395  \\
\Xhline{1pt}
\end{tabular}
    \begin{tablenotes}  
       \item AET: Average Extra Time compared to Centralized in Success cases
     \end{tablenotes}
    \vspace{-0.1in}
    \label{tab:narrowway}
    \end{threeparttable}
\end{table}
The proposed method generated interactions successfully under all random settings. The \textit{vanilla} version has a higher success rate than the \textit{the brake} version but caused more collision during closed-loop simulating. For extra time cost for interaction, we use the closed-loop interaction completion time for the same setting under the centralized potential game setting as the baseline and only calculate the average extra time for success cases. All distributed settings experience increased interaction completion time, but our IPG setting is the most efficient in solving interaction.

An illustrative comparison of interaction using different distributed interaction settings is shown in Figure \ref{fig:distributed comparison} using different stages of interactions. The \textit{vanilla} distributed agents stuck at a deadlock for a long time during interaction. We didn't show the \textit{brake} agents since they experienced deadlock and got stuck from the beginning.
Using IPG, with an imagined game in mind, agents assume the presence of cooperation and predict others' cooperation using the estimated parameters. This results in one agent yielding to the other or assuming the other's yielding in interaction. 

\begin{figure*}[t]
    \centering
    \includegraphics[width=\textwidth]{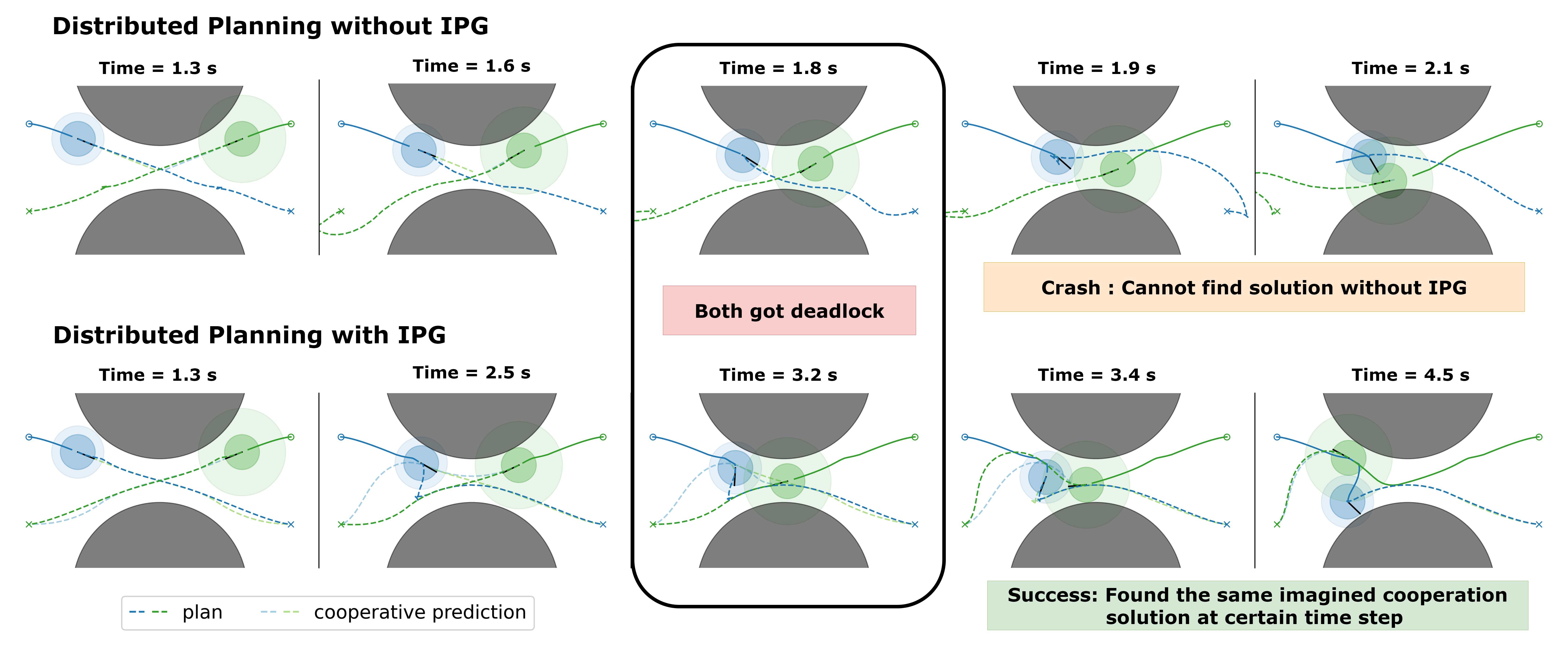}
    \vspace{-0.2in}
    \caption{Vanilla distributed agents can collide during simulation when facing deadlocks. Agents with IPG also experienced deadlock when the imagined cooperation didn't match. However, IPG agents can converge to success plans in the closed-loop simulation.}
    \label{fig:deadlock}
\vspace{-0.25in}
\end{figure*}

We show how IPG recovers from open-loop deadlocks in Figure \ref{fig:deadlock}. In the distributed setting, we don't have a guarantee that agents don't get contradicting open-loop plans causing deadlocks. At 3.2s for the \textit{IPG} setting, the blue agent had wrong predictions on the green agent, causing the deadlock. However, as the interaction progressed, we observed the predictions converging to correct ones. If agents are assigned non-identical safety parameters, they have a low chance of always getting symmetric and contradicting plans.
% To summarize, incorporating an imagined game into planning helps solve the infeasible problem in a distributed setting and efficiently generates a safe and smooth interactive trajectory.

\subsection{Realistic and Diverse Behaviors with IPG}
\noindent\textbf{Realistic Behavior in a T-intersection case}
A typical case and extension for the narrow-way case is the T-intersection case, where the obstacles obscure the agents until one arrives at the intersection. In centralized planning, the optimal cooperative strategy is to wait at the starting point and move until the other agent passes. However, in the distributed setting, the agent cannot observe the other agent until it enters the intersection. Therefore, as shown in Figure \ref{fig:T-intersection}, the agent entered the intersection, realized it had to yield to the other agent, and then retreated to wait. These realistic and human-like reactions happen only in distributed settings. Full animations of interactions and comparisons with baselines for different scenarios can be found on the website\cite{website}. 
In general, evaluating realistic behavior in rich interactions is hard since these interactions only happen with certain initial conditions, and there's no quantitative metric for realism. We found adding imagined cooperation into planning is an effective way to generate realistic behavior, especially in indoor scenarios, where collision-free trajectories are often not available for some agents.

\begin{figure}[h!]
    \centering
    \includegraphics[width=0.5\textwidth]{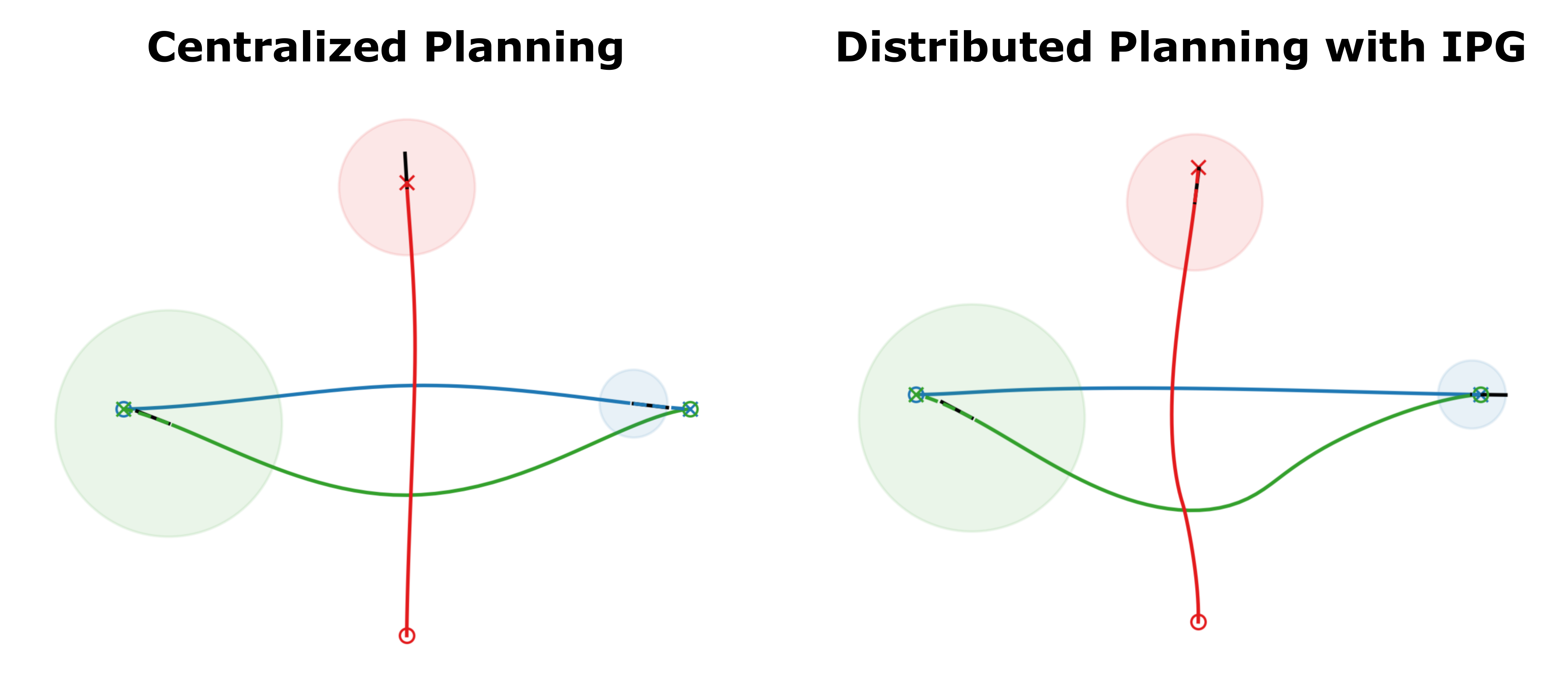}
    \vspace{-0.2in}
    \caption{Three agents interacting with different safety radii.}
    \label{fig:3agent}
\vspace{-0.1in}
\end{figure}

\noindent\textbf{Interactions of more than two agents}
Fig \ref{fig:3agent}. illustrates the distinct behaviors exhibited by three agents in a centralized and distributed IPG setting. To demonstrate the different behaviors of agents with different safety radii, we chose an open area to interact. In the distributed setting, agents lack the safety distance information of others, and they assume that other agents possess equal safety distance. Therefore, the agent with a larger safety distance (green) will react more conservatively.

\begin{figure}[h]
    \centering
    \vspace{-0.05in}
    \includegraphics[width=0.5\textwidth]{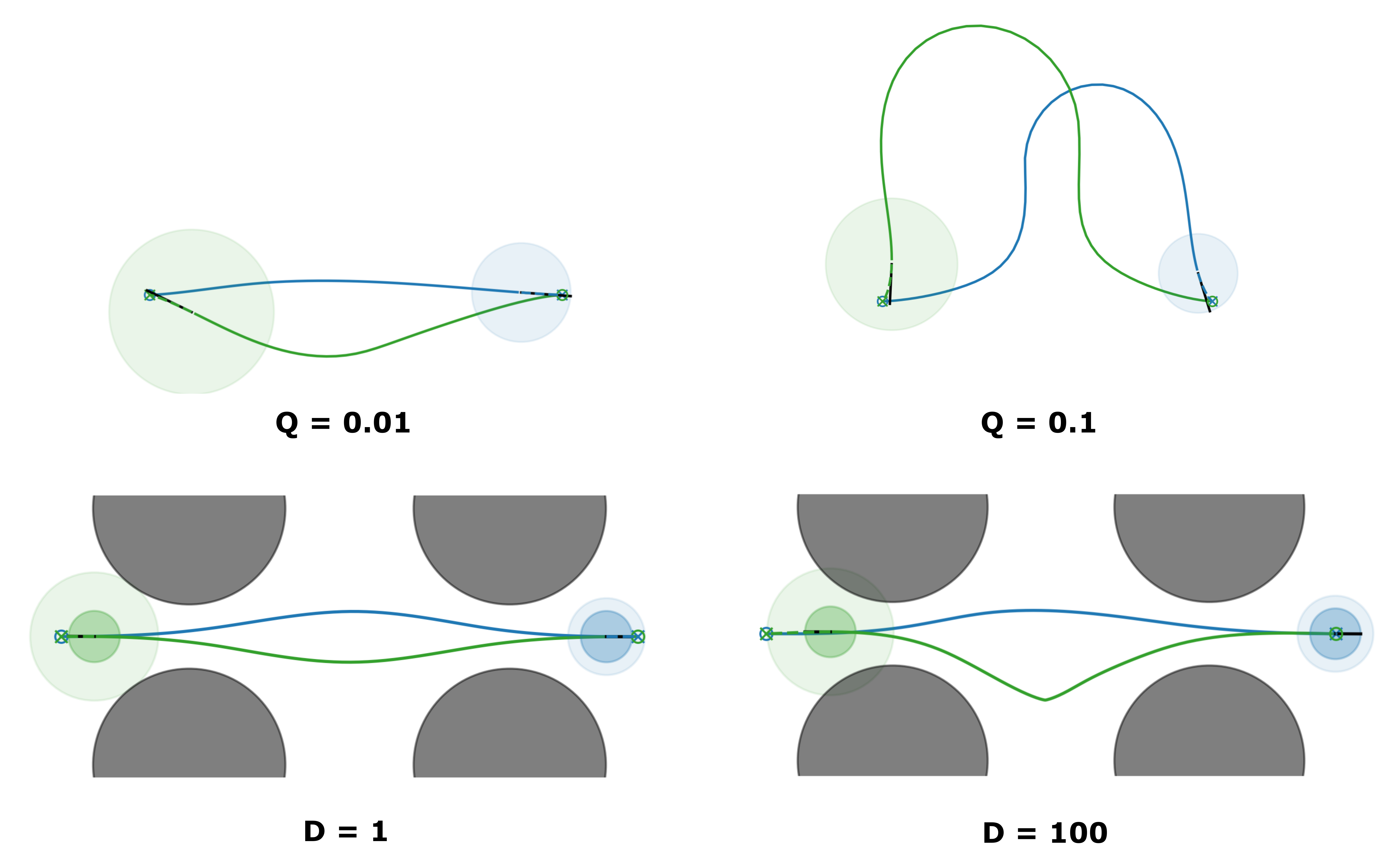}
    \vspace{-0.2in}
    \caption{Effects of different interaction parameters on behaviors.}
    \label{fig:ablations_parameters}
\vspace{-0.15in}
\end{figure}

\noindent\textbf{Effects of different interaction parameters}
The effect of safety and cost parameters on agents' behavior is shown in Figure \ref{fig:ablations_parameters}. The state cost weight $Q$ determines the flexibility in planning, and the safety weight $D$ affects its conservatism in interaction. More examples are shown on the website\cite{website}.

\subsection{Interaction between Heterogeneous agents}
\label{sec:heterogeneous}
In the previous two sections, we've demonstrated the proposed IPG framework for simulating interactions under the distributed setting by preventing collision and deadlocks. An extension of this framework, as mentioned in the introduction, is to use the proposed framework as evaluation agents to interact with tested agents. This requires the IPG agent to interact with different types of agents. 

\begin{figure}[h]
    \centering
    \includegraphics[width=0.5\textwidth]{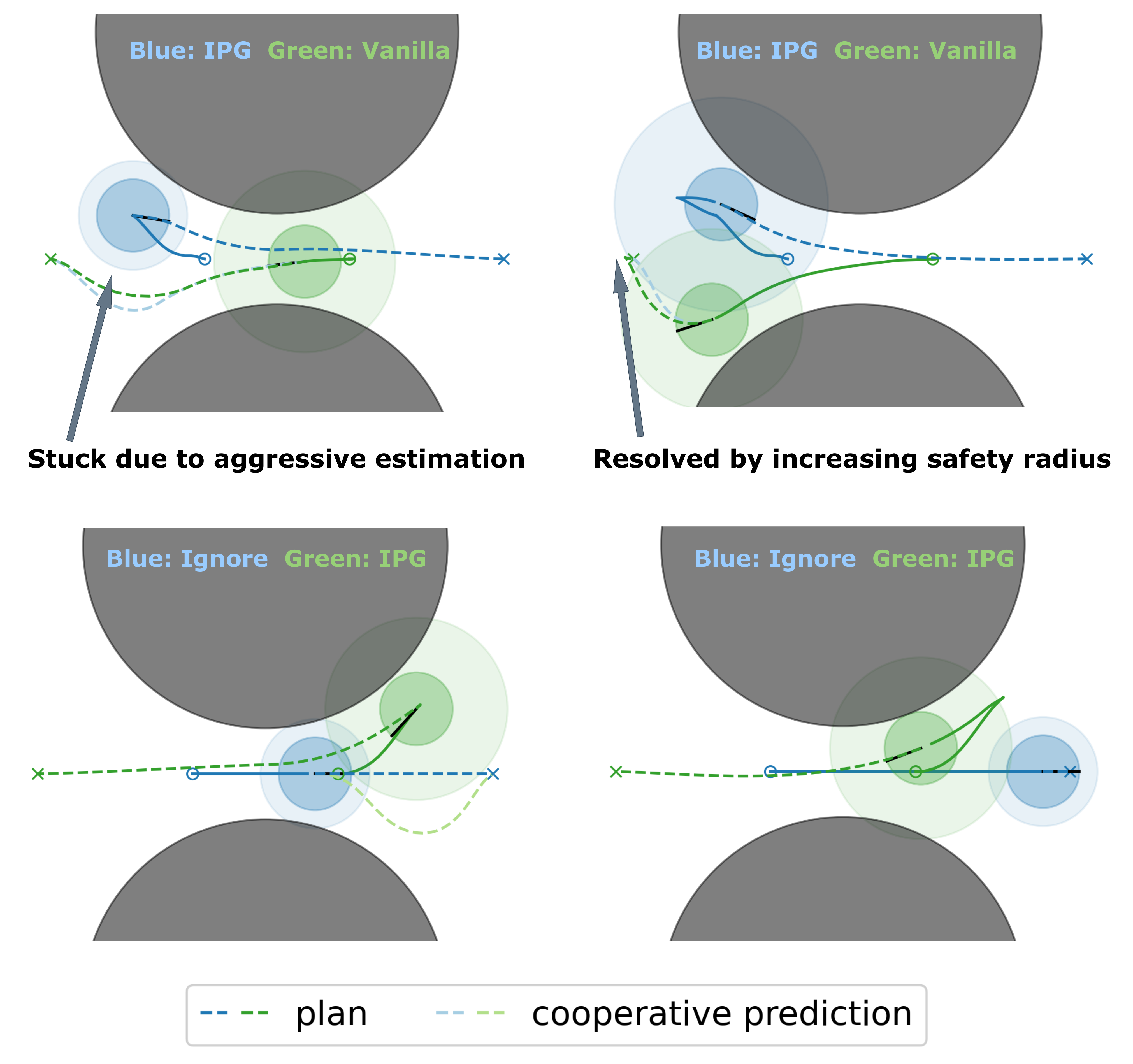}
    \vspace{-0.25in}

    \caption{Interactions between of heterogeneous agents.}
    \label{fig:heterogeneous}
\vspace{-0.2in}
\end{figure}

In Figure \ref{fig:heterogeneous}, we demonstrate the cases where an IPG agent is asked to interact with a \textit{vanilla} agent using constant velocity prediction, and a \textit{ignore} agent ignoring other agents, the IPG agent can react wisely to those non-cooperative agents. 
However, IPG agents with aggressive or non-adaptive parameters might collide or get stuck in interactions. We claim various IPG agents can represent different human agents in the simulator to test the robustness and generalizability of social navigation algorithms. 

\section{Discussion}

\noindent\textbf{Conclusion:}
In this work, we propose an Imagined Potential Game framework under the distributed without communication setting to model realistic interactions in complex scenarios. 
We demonstrate the improvement of IPG agents in the distributed setting in success rate and navigation efficiency in simulating interactions and the ability to generate realistic and diverse interactions in different scenarios.

\noindent\textbf{Limitations and Future works}
The current framework is analyzed in selected scenarios where cooperative interaction is required to solve the problem. In future works, we will adapt the framework for general in-door scenarios where obstacles have random and non-convex shapes. We also plan to develop a test benchmark using IPG agents with various interaction parameters to test different social navigation algorithms.

\bibliographystyle{IEEEtran}
% \bibliography{sample}
\bibliography{reference}

\addtolength{\textheight}{-12cm}   % This command serves to balance the column lengths

\end{document}